%% file: main.tex
\documentclass[twocolumn, switch]{article} 

\usepackage{amsmath, amsthm, amssymb, amsfonts}

\usepackage[numbers,square]{natbib}
\usepackage[utf8]{inputenc}	
\usepackage[T1]{fontenc}	
\usepackage{xcolor}		
\usepackage{hyperref}
\usepackage{multirow}
\usepackage{booktabs} 		
\usepackage{nicefrac}		
\usepackage{microtype}		
\usepackage{lineno}		
\usepackage{float}			
\usepackage{algorithm}
\usepackage{lipsum}		
\usepackage{tabularray}
\usepackage{amsmath,amssymb,amsfonts}
\usepackage[normalem]{ulem}
\usepackage{algpseudocode}
\usepackage{hyperref}

\usepackage{newfloat}
\DeclareFloatingEnvironment[name={Supplementary Figure}]{suppfigure}
\usepackage{sidecap}
\sidecaptionvpos{figure}{c}

\usepackage{titlesec}
\titlespacing\section{0pt}{12pt plus 3pt minus 3pt}{1pt plus 1pt minus 1pt}
\titlespacing\subsection{0pt}{10pt plus 3pt minus 3pt}{1pt plus 1pt minus 1pt}
\titlespacing\subsubsection{0pt}{8pt plus 3pt minus 3pt}{1pt plus 1pt minus 1pt}

\usepackage{tikz,xcolor,hyperref}

\definecolor{lime}{HTML}{A6CE39}
\DeclareRobustCommand{\orcidicon}{
	\begin{tikzpicture}
	\draw[lime, fill=lime] (0,0) 
	circle [radius=0.16] 
	node[white] {{\fontfamily{qag}\selectfont \tiny ID}};
	\draw[white, fill=white] (-0.0625,0.095) 
	circle [radius=0.007];
	\end{tikzpicture}
	\hspace{-2mm}
}
\foreach \x in {A, ..., Z}{\expandafter\xdef\csname orcid\x\endcsname{\noexpand\href{https://orcid.org/\csname orcidauthor\x\endcsname}
			{\noexpand\orcidicon}}
}

\title{Enhancing Screen Time Identification in Children with a Multi-View Vision Language Model and Screen Time Tracker}

\usepackage{authblk}

\author[1, $^{\dag}$]{Xinlong Hou}
\author[2, $^{\dag}$]{Sen Shen}
\author[1, $^{\dag}$]{Xueshen Li}
\author[3]{Xinran Gao}
\author[4]{Ziyi Huang}
\author[5]{Steven J. Holiday}
\author[6]{Matthew R. Cribbet}
\author[6]{Susan W. White}
\author[7]{Edward Sazonov}
\author[1,*]{Yu Gan}

\affil[1]{Department of Biomedical Engineering, Stevens Institute of Technology}
\affil[2]{Department of Computer Science, Iowa State University}
\affil[3]{Department of Electrical Engineering, Columbia University in the City of New York}
\affil[4]{Nokia Bell Labs}
\affil[5]{Department of Communication \& Information Science, University of Alabama}
\affil[6]{Department of Psychology, University of Alabama}
\affil[7]{Department of Electrical and Computer Engineering, University of Alabama}

\begin{document}

\twocolumn[ 
  
\maketitle

\begin{abstract}
Being able to accurately monitor the screen exposure of young children is important for research on phenomena linked to screen use such as childhood obesity, physical activity, and social interaction. Most existing studies rely upon self-report or manual measures from bulky wearable sensors, thus lacking efficiency and accuracy in capturing quantitative screen exposure data. In this work, we developed a novel sensor informatics framework that utilizes egocentric images from a wearable sensor, termed the screen time tracker (STT), and a vision language model (VLM). In particular, we devised a multi-view VLM that takes multiple views from egocentric image sequences and interprets screen exposure dynamically. We validated our approach by using a dataset of children's free-living activities, demonstrating significant improvement over existing methods in plain vision language models and object detection models. Results supported the promise of this monitoring approach, which could optimize behavioral research on screen exposure in children’s naturalistic settings.
\end{abstract}

\vspace{0.35cm}

] 
\footnotetext{Co-first authors$^{\dag}$. Corresponding authors$^{*}$: \url{ygan5@stevens.edu}}

\input{chapters/introduction}
\input{chapters/related_work}

\input{chapters/method}

\input{chapters/exp}

\section{Discussion}
In this study, we propose MV-VLM, the first known vision-language model that is deployed for screen type identification.  The dual extraction performance distinctly surpasses that of other existing models. The efficacy of MV-VLM in processing and integrating these diverse data types indicates potential utility in applications where comprehensive understanding from multiple perspectives is crucial. Moreover, MV-VLM only requires a minimal amount of training samples, as the training is on the alignment layers and the other models are fine-tuned on a pre-trained models (Swin Transformer and MiniLM). Such feature enhances the feasibility of deploying our model in resource-constrained environments. Besides the data analysis, we also improve the data collection process. Unlike previous studies that collected data from adults in controlled environments, this study focuses on data collected from children in a free-living environments, making the dataset more representative of real-world usage and future behavioral studies. To protect children's privacy, guardians and children may delete images that they feel uncomfortable at the end of data acquisition following conventional protocols in \cite{hassan2020selective, kelly2013ethical}.

There are a few limitations of this study. First, we only examine the case of single screen type is involved. We will expand the framework to multi-task classification when multiple screens are involved in sequential images. 
Second, our data collection has been restricted to indoor environments, which limits the exposure of our model to varied environmental contexts. In future research, we plan to expand our data acquisition efforts to outdoor settings, thereby enriching the model’s training dataset with a broader range of environmental dynamics. This expansion is anticipated to booster the model’s generalization capabilities and enhance its performance across more diverse real-world scenarios.

To ensure a fair evaluation on the algorithm development, we exclude samples with severe motion blurring and images accidentally covered by clothes or other objects. We consider those images as outliers which do not correspond to features captured by wearable sensors. In the integration of a hardware system, we plan to use accelerometers to inform the exclusion of low quality blurry images and use ambient light sensors to instruct the system that there are occlusion and the Multi-View images are not needed. Future study will seek to integrate behavioral research to measurement obtained from MV-VLM. By correlating our model’s outputs with behavioral data, we aim to deepen our understanding of the interactions between environmental contexts and individual behaviors. This approach is promising not only to validate MV-VLM but also to extend to more complex, behaviorally-driven studies. Moreover, in this study, we focused on identifying single screen types. However, our framework naturally supports the identification of multiple screens within a scene. In the future, we will explore the application of identifying multiple screen types. Moreover, we will consider the proximity of the screen in our framework by leveraging the depth information \cite{depthanything}, further improving the performance of the proposed framework in the future.

\section{Conclusion}
We proposed a lightweight and comfortable wearable sensor solution using screen time tracker to monitor children's screen exposure. We devised a Multi-View vision language model to identify existence of screens from egocentric image streams. We collected data from children's free living activities. We explored Multi-View images to enhance features of screen identification. This Multi-View model is integrated with vision language model, generating image descriptions that are related to screen existence. Our experiments indicate superiority in comparison with conventional vision language model and object detection model. Future work will include accumulative measurements of screen time exposure and associate the screen measurements with other factors in behavioral studies.

\section*{Acknowledgments}
Research reported in this study was supported by Institute of the National Institutes of Health under award number R21HD104164. The authors would like to thank Olivia Castro for the assistance in annotation.

\bibliography{mybibliography}{}

\end{document}

%% file: chapters/introduction.tex
\section{Introduction}

Screen exposure has garnered an increasing attention in the past decades due to the dramatic rise in the digital technology. Extensive screen exposure has been associated with healthy and psychological problems \cite{aishworiya2022there} such as eye problems \cite{Yang.2020}, language disorders \cite{Collet.2019}, sleep disorders \cite{Hale.2015}, obesity \cite{Robinson.2017b}, and cognitive impairments \cite{Supanitayanon.2020}. In particular, following the suggestions from World Health Organization (WHO), children with age between 2 and 4 should have less than 1 hour of screen exposure time per day \cite{world2019guidelines}. However, this age group has been reported to spends an average of 2.5 hours in front of screens\cite{viner2019health}.
Therefore, it is important for parents to objectively monitor children's screen exposure and manage children's screen activity. In addition, there is also an unmet need for scientists to accurately measure screen time to better understand the association between health concerns and screen exposure \cite{perez2023validated}. Existing research methods for measuring screen exposure rely on users' self-reporting, experimental technologies (e.g., eye-tracking glasses), or built-in device apps. Whereas self-reporting is prone to bias, eye-tracking glasses can be invasive, and built-in apps can't measure exposure across devices or confirm users' identity. There is currently no non-invasive and automatic solution for accurate and robust measurement of screen exposure on cross-device screens.


\begin{figure}[!t]
\includegraphics[width=0.48\textwidth]{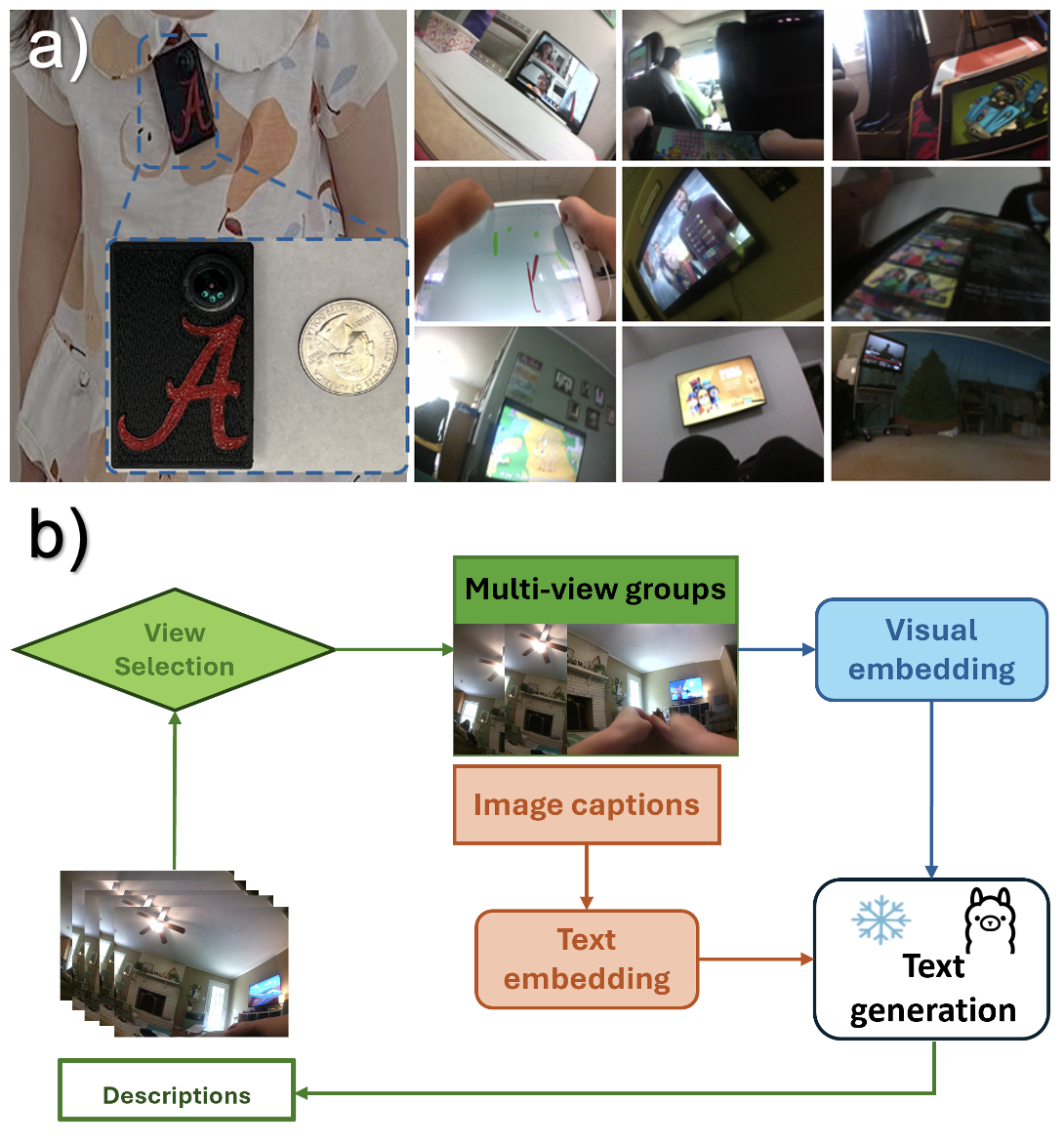}
\caption{a) The STT device (left panel); Compilation of images showcasing various environments in screen time exposure. The montage is created using free-living data collected from STT (right panel). The STT device is lightweight and can be firmly attached to clothes. b) The architecture diagram of the proposed MV-VLM. The MV-VLM is designed to process egocentric camera frames collected by a).}
\label{device}
\end{figure}

A wearable camera with the capability to capture egocentric images is a promising non-invasive candidate to monitor screen presence. Egocentric images are captured from first-person perspective and can demonstrate the wearer's viewpoint. Egocentric images have been utilized for human activity analysis \cite{xu2023towards, liu2021enhanced,huang2021holographic}. 
Although there is existing work that uses wearable cameras to track children’s screen exposure \cite{Downinge028265, everson2019can}, those solutions are not ideal due to their bulky size and lack of automated analysis. Artificial intelligence (AI) enables computers to automatically identify objects with high accuracy, making it promising for screen exposure measurement. While AI can identify object types from egocentric images, no automated algorithm has been customized for children's screen exposure. 
Recently, vision language models have shown great potential for robustly identifying objects and events from images \cite{zhang2024vision, 10243119, 10149530}. 
However, no such efforts have been made specifically for egocentric images for screen type identification. Convolutional neural network (CNN), as a major branch of AI, has advanced automated object detection methods in various applications by providing the backbone architectures for the development of the regional based CNN (R-CNN)\cite{ Girshick_2015_ICCV} and You Only Look Once (YOLO)\cite{AlexeyBochkovskiy.2020} detection systems. However, such methods work on individual frames and relies on local features extracted from CNN, lacking the capability to associate long-range dependency of features or high-level objects.

In this paper, we proposed a combination of an egocentric camera, namely the screen time tracker (STT), and a vision language model (VLM) to identify screen existence among multiple screen devices. The STT device and representative images are shown in Fig. \ref{device} a). In particular, we devised a multi-view VLM (MV-VLM) to process sequential frames from egocentric images as shown in Fig. \ref{device} b). Large language models provide a human-readable explanation to the insights of images compared with traditional classification methods. In addition, it unveils the spatial relationship among multiple objects that are relevant to the existence of the screen. Particularly, with the involvement of transformers in LLM, long-range dependence of local features is taken into consideration. Moreover, the multi-view scheme also explores the temporal features, making our proposed model particularly capable of detecting a screen that is in occlusion. We took advantage of a customized wearable sensor and egocentric images to identify children's screen exposure. In comparison with existing single view input vision language models, we developed a unique multi-view vision language model that process egocentric image stream from wearable devices. Notably, a contrastive learning-based view selection module and a screen type identification module are specifically designed to address the challenges in existing method and to develop a robust approach to detecting children’s screen exposure. The proposed approach innovatively takes multi-view rather than single-view images as input compared to the existing screen identification work \cite{kelly2013ethical, hassan2020selective}. The innovation contributes to the model's capability in extracting long-dependent textual features alongside spatial-temporal features from multi-view imaging, achieving the highest performance among all three types of screen scenarios.

Our major contributions are summarized as follows:
\begin{itemize}
    \item We proposed a wearable sensor solution to identifying children's screen exposure. Our lightweight wearable sensor is children-friendly and captures egocentric images to monitor children's electronic screen activities. We created a dataset collected from children's free-living activities over two days, the first of its kind for egocentric camera.
    
    \item We designed the first vision language model to identify children's screen time using wearable sensors. This model processes egocentric images to generate descriptive content, from which screen type identification is achieved through keyword extraction from the generated text.
    
    \item We devised a MV-VLM model, which is specifically for analyzing image stream from wearable cameras. To leverage temporal variations and capture more spatial features for accurate type identification, we developed a novel model that takes multi-view inputs and fuses features from different views for language generation. The selection of multi-view images is based on contrastive learning, which maximizes the information among images from different views. 
\end{itemize}

%% file: chapters/related_work.tex
\section{Related Work}

\textbf{Screen Exposure and Wearable Cameras.} Understanding the link between health issues and screen exposure is crucial, necessitating precise tracking of screen usage. Several studies have explored the extent of children's screen exposure using wearable cameras, which are non-invasive and less dependent on bias compared to self-reporting \cite{su2024assessing, sadeghian2022first}. Wearable cameras also indicate a majority distribution of screen exposure goes to TV (42.4\%) of children’s daily screen exposure in \cite{lowe2023watching}. Overall, very limited work \cite{su2024assessing} has explored the solution to automate the screen detection from wearable cameras.

\textbf{Traditional Neural Networks in Egocentric Videos.} However, while many of these works rely on manual annotations, an automated solution for data analysis could largely reduce human workload. In AI field, deep learning models have been largely adapted in processing the videos from the egocentric wearable camera. Chen et al. \cite{chen2021recognizing} utilizes CNN and random decision forest for activity recognition; Song et al \cite{song2024video} utilizes CNN-based and VLP-based models in extracting surrounding information for visually impaired person;  Bock et al\cite{bockwear} utilizes ActionFormer for outdoor sports recognition;\cite{qiu2021indoor} utilizes an LSTM-based encoder-decoder framework to predict movement trajectory of a targeted person.  The integration of multi-modal data such as inertial data \cite{bockwear} and IMU readings \cite{qiu2021indoor} demonstrated an improved performance compared to using vision-based images alone.

\textbf{Multi-view based Neutral Networks.} A multi-view deep learning model leverages data from multiple perspectives to enhance learning in tasks such as classification and recognition by integrating diverse sources of information. Its potential has been explored in recent research \cite{9712308, 10415089}. Although it has been demonstrated in existing studies that the incorporation of multi-view in CNN-based models outperforms models with single-view \cite{8202213, 7926630}, multi-view network has not been fully explored in processing egocentric image stream. A critical problem in applying multi-view network is how to select multi-view images from egocentric image stream.

\textbf{Large language model (LLM) in Egocentric Streams}. Large language model is a rising field and incorporating LLMs into image processing has the potential to significantly improve upon the studies using traditional models such as CNNs. By leveraging the contextual capabilities of LLMs, researchers can achieve more accurate and comprehensive egocentric image understanding.  LifelongMemory was proposed in \cite{wang2023lifelongmemory} as a novel framework that uses multiple pre-trained models to  answer queries from egocentric video content. Research in \cite{wang2023egocentric} addresses Ego4D natural language queries challenge with image and video captioning models.  Most of the studies focuses on answer queries, thus lacking the capability to systematically analyze daily life for specific applications. Moreover, there is limited vision language model to address the identification of electronic screen type, even from a single view. 

\begin{figure*}[!t]
\centering
\includegraphics[width= 0.94\textwidth]{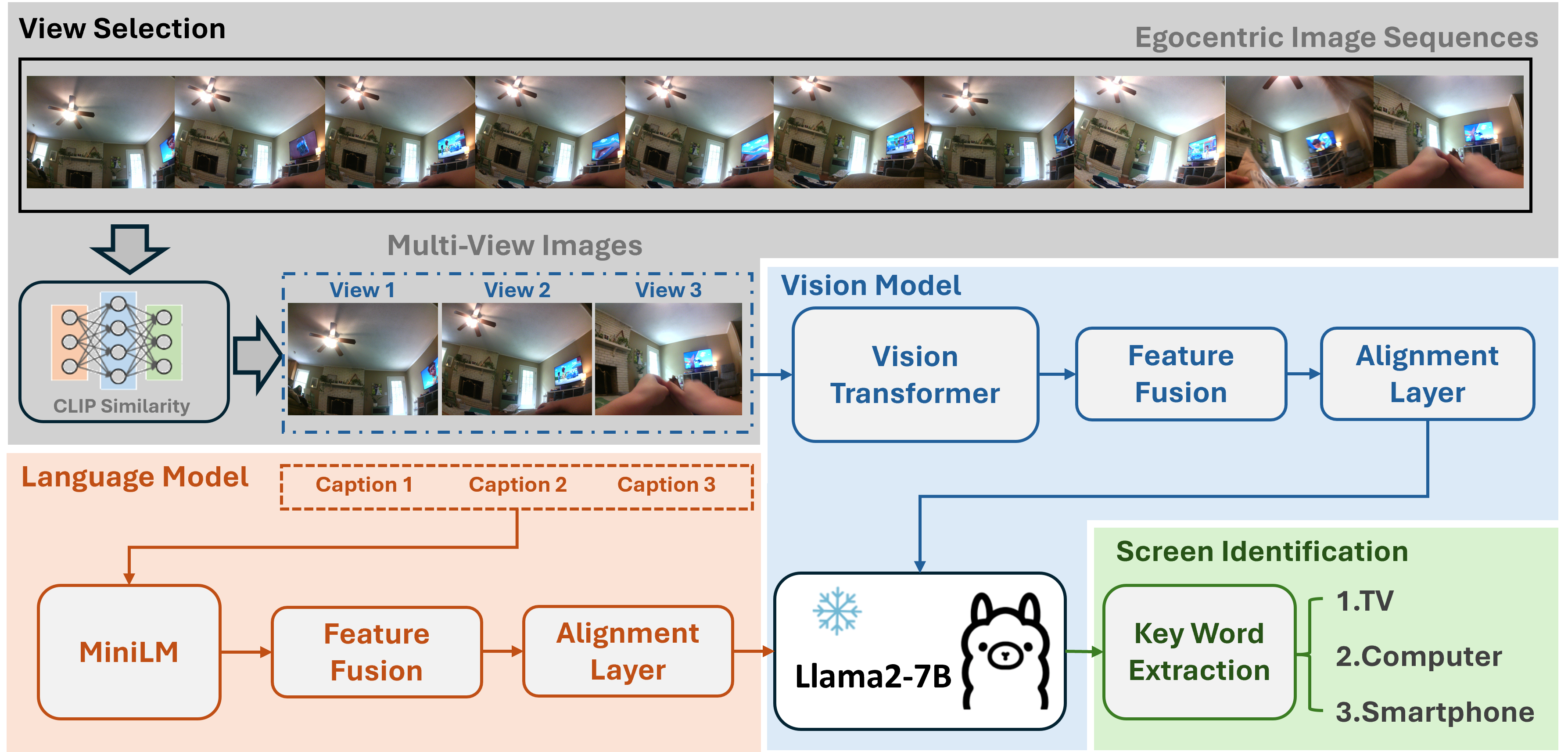}
\caption{Proposed pipeline to process egocentric image stream. There are four major components: view selection, vision language model, language model, and screen identification. View selection module uses Contrastive Language-Image Pre-Training (CLIP) to extract embeddings and select multi-view i   mages based on similarity. Vision Langauge model learns from vision transformer and MiniLM to generate textual description on multi-view images. }
\label{flow_chart}
\end{figure*}

%% file: chapters/method.tex
\section{Method}

\subsection{Wearable sensor and data collection}
We collected data from an egocentric device, screen time tracker (STT), following our previous work \cite{li2022non}. The key component used in the sensor was a miniature 5 Megapixel camera with 120-degree wide-angle gaze-aligned lens with the resolution of $2592 \times 1944$. The camera was attached to clothing with a magnetic clip mount, as shown in Fig. \ref{device} a) (left panel), and boasted a battery life exceeding 48 hours. We redesigned the device's casing by resembling a badge to reduce user burden, particularly for children. Positioned on the chest, this badge-like device captured egocentric images every 10 seconds. Our device measured a dimension of $54 mm \times 35 mm \times 12 mm$. Notably, the device only takes 30\% of the size of commercial cameras used in other wearable senor informatics work \cite{Downinge028265}.

Young children between age of 3 and 5 were eligible to participate in data acquisition. The experimental protocol was approved by the local Institutional Review Board (IRB) at home institution. We collected data from 30 participants. Each participant wore the device for two days. The whole dataset roughly includes children's participation in free-living activities. After the two day activities, a survey on the comfort of the wearable device is conducted following a conventional protocol \cite{hassan2020selective, kelly2013ethical}. The screen exposures related to TV, computers, and smartphones were captured. Representative images are shown in Fig. \ref{device} a) (right panel). Image data were manually labeled by an undergraduate research assistant.






    

\subsection{Framework}

Our framework collected multiple images from different views to identify the existence of electronic screens. As shown in Fig. \ref{flow_chart}, we developed a novel screen identification framework that consists of a view selection module, a vision model, a language model, and a screen identification module. The core concept involves selecting and inputting images from multiple egocentric viewpoints to enhance the robustness of electronic screen identification.
The language model unveils the spatial relationship among multiple objects that are relevant to the existence of the screen. Particularly, with the involvement of transformers in LLM, long-range dependence of local features is highlighted. The Multi-View scheme also explores the temporal features, making our proposed model particularly capable of detection screen that is in occlusion.

\subsubsection{Conceptual Rationale for Multi-View Vision Language Model} As shown in Fig. \ref{flow_chart}, due to the motion of children during an screen event, the egocentric image stream usually includes views at various angles to the electronic screen.
In the screen exposure scenarios, Multi-View image processing could be beneficial in several aspects. For instance, when screens are only partially captured, Multi-View images may provide complementary data from different parts of the screen, enabling deep learning models to better understand and integrate the screen's various structural components. Additionally, in cases of low-quality images, such as those that are blurry, it is challenging for deep learning models to classify the images. Utilizing Multi-View images allows the model to synthesize features from various perspectives, thereby enhancing feature representation and improving classification accuracy.
To identify the existence of electronic screens, it is essential to consider the spatial relationships among static objects like TVs, computers, windows, walls, and shelves. These objects maintain consistent spatial relationships in the real world, which are reflected in 2D images based on perspective principles. To capitalize on these features and their spatial interrelations, we employed a vision transformer coupled with a language model. This approach leveraged the model's capability to explore long-range dependencies in text descriptions and image processing.

\subsubsection{View Selection}
The image stream captured from STT is in time series, representing objects from different views and timestamps. 
The term 'view' used in this paper is similar to 'perspective'. It refers to images captured from the same camera at different time instants, but within the same scene. This view selection module converts temporal change in images frames to spatial change in perspective/view. Images taken from the same scene with small variations of position, height, and orientation are considered similar, though environmental condition like light, time of the day may change. We sought for a mapping rule that is robust enough to reveal spatial relationships and partial or occluded screen information. To capture coherent image features, we chose a contrastive learning-based embedding approach to encode the images and analyze their similarity. Assuming the egocentric image stream $I_1, I_2, I_3, ..., I_n$, for each image $I_i$, we used contrastive language-image pre-training (CLIP) \cite{radford2021learningtransferablevisualmodels} to convert images to embeddings $CLIP(I_i)$. 
The CLIP enables pairwise image similarity comparison without prior knowledge, making it more adaptable than heuristic methods such as K-Means, which rely on predefined clusters and struggle with the variability of real-world screen detection images.

For any two images $I_m$ and $I_n$, we measured the cosine similarity between two embeddings via:
\begin{equation}
Sim(I_{m}, I_{n})=\frac{CLIP({I_{m})}\cdot CLIP({I_{n}})}{||CLIP({I_{m})}||||CLIP({I_{n})}||}
\end{equation}
Where ($\cdot$) represents dot product and $||\cdot||$ is the magnitude. Then to split the image stream into Multi-View image groups, we built a graph $G(V,E)$ to represent the images with their similarity according to values of $Sim(I_{m}, I_{n})$. In Graph $G$, each node $v_i$ represents an image $I_i$. $V$ corresponds to the set of nodes (i.e., images) and $E$ corresponds to the set of weighted edges that connect the each pair of nodes within a time window. For two nodes $v_m$ and $v_n$, the edge $E_{m, n}$ of them is $valid$ only if $Sim(I_{m}, I_{n})$ is within a range between an upper bound $\tau_h$ and a lower bound $\tau_l$. The upper bound ensured that static images in consecutive frames would not be selected. The lower bound ensured these two images could capture the same screen-related scene. We built an undirected graph and selected connected components of size $k$ as Multi-View image groups. To get the most number of Multi-View groups, we first sorted the all connected components of size $k$ according to the sum of edge degrees in each $k$-size component ascendingly. Then we greedily selected and split the components starting from the components with least edge degrees from $G$, and updated $G$ accordingly. The detailed procedures are illustrated in Algorithm \ref{alg}.



\begin{algorithm}[t!]
  \caption{View Selection}
  \begin{algorithmic}[0]
   \State \textbf{Input}: Graph $G(V,E)$
   \State \textbf{Output}: Set of Multi-View image groups ($S$)
   \State \textbf{Parameters}: Valid components ($S_v$);
   
    \State $S_{v},S \gets []$
    \For{$E_{i,j} \in E$}
        \If{$E_{i,j}$ is $not$ $valid$}
            \State Delete $E_{i,j}$ from $E$
        \EndIf
    \EndFor
    
    \For{Connected component $g \subseteq G$}
        \State Append $g$ to $S_v$
    \EndFor
    \State Sort $S_v$
    \For{Connected component $g \in S_v$}
        \If{$g \subseteq G$}
            \State Append $g$ to $S$
            \State Delete $g$ from $G$
        \EndIf
    \EndFor
  \end{algorithmic}
  \label{alg}
\end{algorithm}

\subsubsection{Vision Language Model}
The VLM model includes three components: a ViT model for vision embedding, a MiniLM model for text processing, and a Llama model for text generation. The three modules are connected by alignment layers.

\textbf{Visual Embedding (ViT)}. We utilized the Swin Transformer \cite{9710580} for image embedding. The transformer model used in this paper is a pre-trained model on a conclusive natural image dataset \cite{deng2009imagenet}. Images were divided into smaller non-overlapping patches and learnt by the transformer blocks. The transformer blocks operated in a hierarchical manner and each block applied multi-head self-attention mechanisms based on shifted windows. These shifted windows enabled a set of self-attention models to dynamically learn and extract image features which include long-range dependence. In our model, the features extracted from different views were embedded to a fixed length and sent to fusion layer. In the feature fusion layer, features with the same length were normalized and stacked. Then an alignment layer, which was built by a fully connected layer, fused them into a constant dimension. In particular, the features were averaged to create an input for the following vision-language generating task. This process fused Multi-View images equally to enhance a complementary integration of features extracted from Swin Transformer. 

\textbf{Text Embedding (MiniLM)}. In the training phase, we employed a MiniLM \cite{10.5555/3495724.3496209} for extracting textual features from annotations associated with Multi-View images. In particular, each images were associated with a sentence of textual annotation during training. MiniLM is made of a set of distilled self-attention model in transformer implementation. In this study, our analysis was constrained by a limited sample space, as all egocentric images were sourced from indoor home environments. Furthermore, the sample set mainly comprises screen-related instances. In addition, textual information involved in this study was mainly that relevant to screen events. This made MiniLM particularly well-suited for our application, as its distilled architecture efficiently captured the necessary textual embeddings during the training phase. In the Feature Synthesis layer, text embeddings were normalized and concatenated to ensure the brevity of image captions. These concatenated embeddings were then processed through a linear alignment layer, ensuring dimensional consistency with the embeddings used in the Swin Transformer model. In particular, the text embeddings extracted from this phase serve as the annotation for the following text generation model.
We adopt the MiniLM for processing the text input for two reasons:
MiniLM maps input texts of varying lengths into fixed-length embeddings, enhancing the robustness and scalability of the proposed framework.
Moreover, by using the embeddings as input, the proposed framework can focus on high-level semantic information from the text.

\begin{figure*}[t]
\centering
\includegraphics[width=0.95\textwidth]{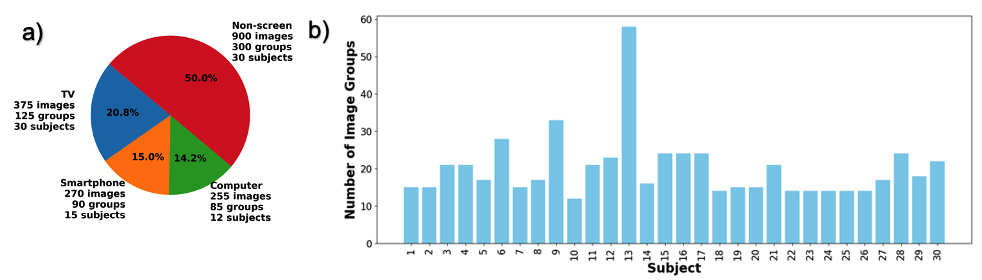}
\caption{Details of dataset acquired in this research. a) The overall distribution of different screen types. b) The number of image groups acquired from each subject.}
\label{fig:distribution}
\end{figure*}

\textbf{Alignment layer.}
The alignment layer is a fully-connected layer which projects embeddings of multi-modal data to the LLMs' input space. The fused multi-modal data by concatenation is sent to the alignment layer. Then processed embeddings are sent to the LLM for screen identification. The alignment layer has trainable parameters that can be updated during the training process, in order to achieve the optimal projection of the multi-modal data to the LLM's input space.






    

\textbf{Text Generation (Llama)}.  In this study, we used a large language model, $Llama2-7B$ \cite{touvron2023llama}, to efficiently produce fast scene description regarding the contents in the Multi-View images. We used one set of fully connected layers, which served as a soft prompt, to align the visual features with the large language model. Similarly, another set of alignment layers connected the text embeddings with the large language model. The training of text generation model was an optimization process for alignment layers, where $\theta_v$ and $\theta_t$ corresponded to the weights in alignment for vision model and the weights in alignment layers for text mode. The loss function for the report generation task was jointly optimized by the combination of loss terms from each task: 
\begin{equation}
\begin{split}
    L_{report}(\theta_v, \theta_t;X_v, X_t,X_p, X_r)= \\
    -\sum^{M}_{l=1}logp_{\theta_v, \theta_t}(x_l;X_v,X_t,X_p,X_{r,<l})
\end{split}
\end{equation}
where $x_l$ is a variable related to the predicted token, $M$ represents the length of the generated text, $X_r$ represents the current prediction text, $X_v$ represents the visual embedding, $X_t$ represents the textual inputs, $X_p$ represents prompts, and $X_{r<l}$ represents the token before the predicted token. This loss function considered textual input, visual input, and token generated, thus seeking an optimized parameter setting to generate reliable scene descriptions.

\begin{figure*}[!t]
\centering
\includegraphics[width=1\textwidth]{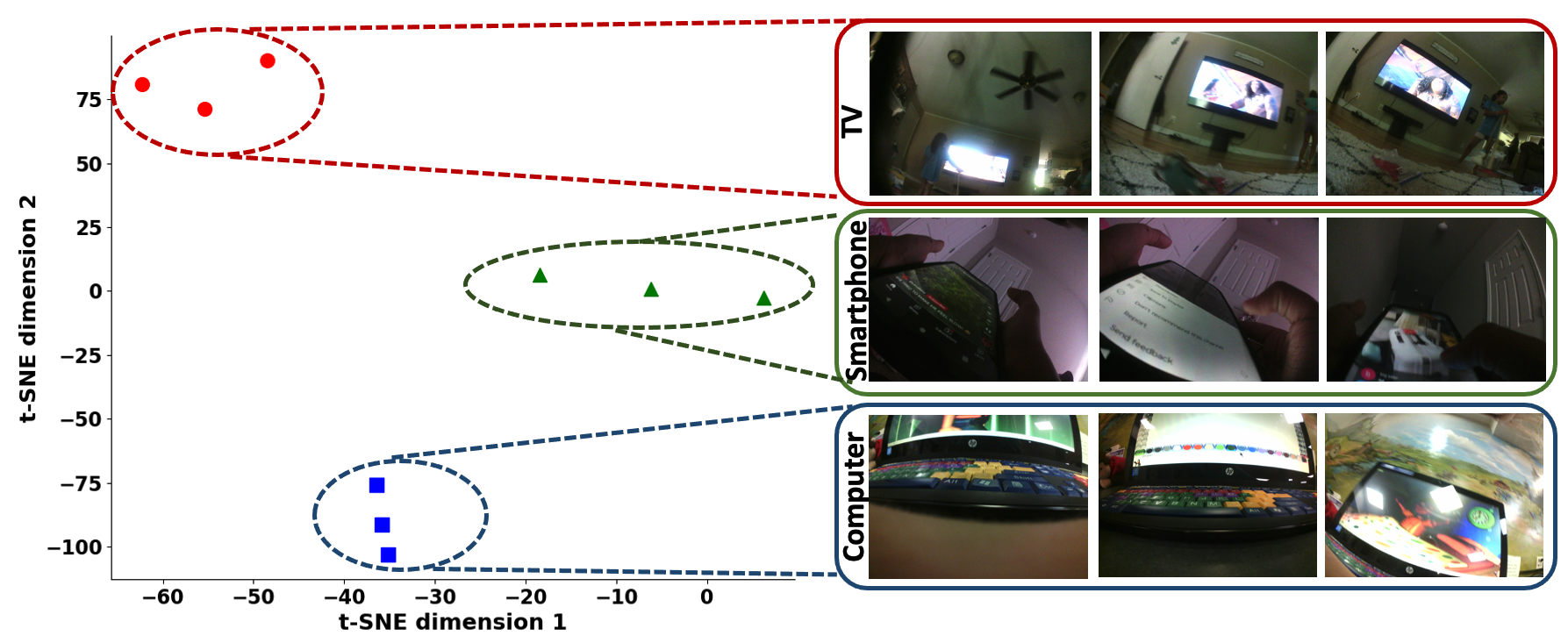}
\caption{Representative Multi-View images selected by CLIP embedding. Our selection maximize the variations among consecutive frames while capturing complementary features of the screen object from different views. The features are visualized by t-SNE.}
\label{Multi-View}
\end{figure*}

\begin{table}[t]
\centering
\caption{A list of mapping between key words and screen types}
\renewcommand{\arraystretch}{1.25}
\scalebox{0.8}{
\begin{tabular}{ll}
\hline
\multicolumn{1}{c}{\multirow{2}{*}{Screen Types}} & \multirow{2}{*}{Key words}                \\
\multicolumn{1}{c}{}                             &                                           \\ \hline
TV                                               & TV, television                            \\
Smartphone                                       & Smartphone, Phone, Tablet, Cellphone, iPad \\
Computer                                         & Computer, Laptop, Computer Monitor                 \\ \hline
\end{tabular}}
\label{table:map}
\end{table}

\subsubsection{Screen Type Identification}
The text description of the Multi-View images was further processed to identify screen types. First, we extracted key words from generated description. The key words included a whole set of screen-related objects, such as "TV", "Television", "Cellphone", "Smartphone", "Monitor", "Computer", "Laptop", etc. Second, key words were categorized to major types through a look-up table, as shown in Table \ref{table:map}. Key words that fell into the same categories were combined and identified as the same screen type. For example, "Smartphone" and "Cellphone" were considered as "Smartphone". "Computer Monitor" and "Laptop" were both considered as "Computer". Due to the similarity between hand interaction with a screen, we combined tablet and smartphone to the same category.

%% file: chapters/exp.tex
\section{Experiments and Results}

\subsection{Dataset}
The dataset has 1,800 images, corresponding to 600 groups of Multi-View images, including different screen types of TV, smartphone, computer use in free-living environments. The condition of the free-living environment includes scenarios where light is bright or dark. The detailed distribution of the screen types is shown in Fig. \ref{fig:distribution}, which aligns with the screen usage distribution in the study \cite{lowe2023watching}. Moreover, we provide the number of image group acquired from each subject. To build a Multi-View image dataset with caption describing the screen of each group, labels were manually generated with guidance from Bootstrapping Language-Image Pre-training (BLIP2) \cite{li2023blip} and Indoor Scene Understanding \cite{wang2024rootvlmbasedindoor}. The captions will provide environment information besides screens. The spatial relationship between screens and the environments will be provided and extracted by the language model for screen detection. We divided the whole image set to four folds based on random shuffle over candidates and conducted cross-validation to validate the performance of identifying various screen types.

\begin{figure*}[!t]
\centering
\includegraphics[width=1\textwidth]{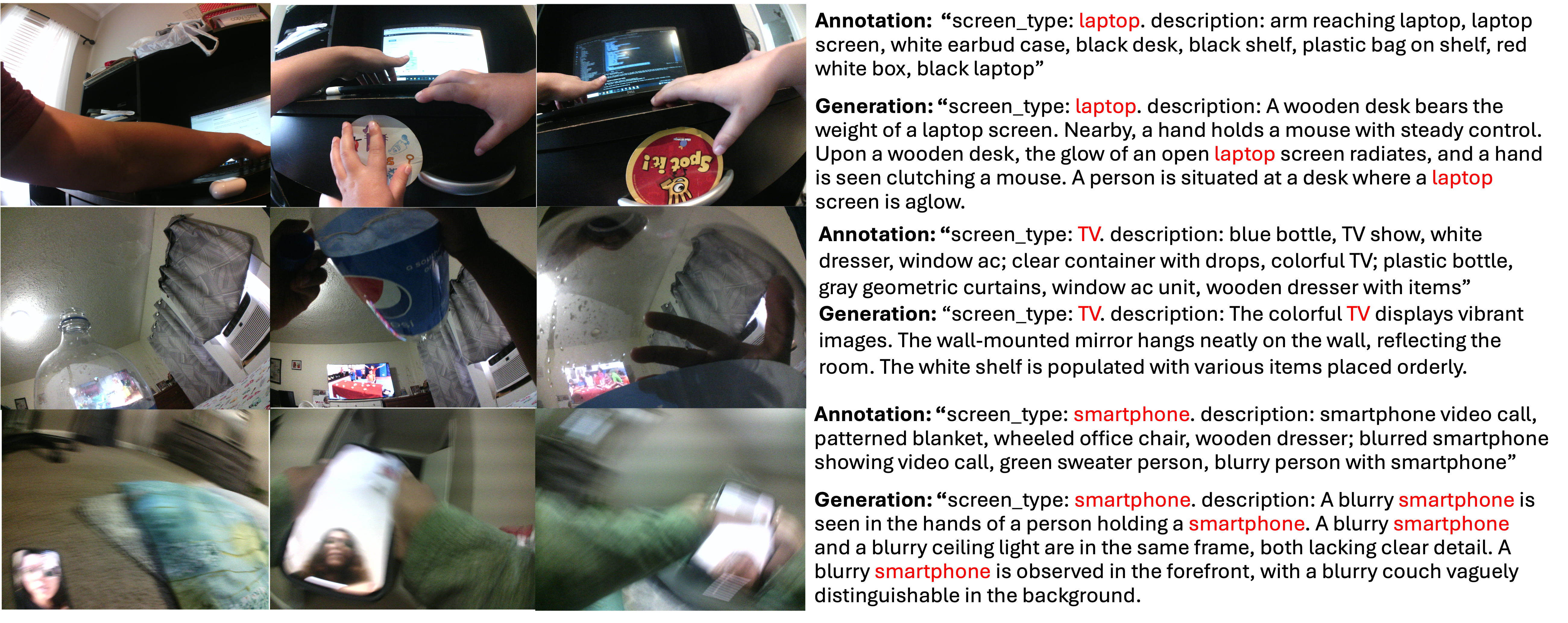}
\caption{Examples of generated text and screen identification. The left panel shows typical Multi-View images. The right panel shows the generated description from language model, the screen identification results, and the annotation. Key words could be efficiently processed to categorize to specific screen type.}
\label{fig:text_res}
\end{figure*}

\subsection{Training strategy for implementation}
The training process required minimal effort to customize the VLM for screen detection. In particular, we chose a pre-trained model in Swin Transformer for ViT \cite{9710580} and MiniLM for language model \cite{10.5555/3495724.3496209}. We fine-tuned ViT using our Multi-View image datasets. In text generation, Llama model ($Llama2-7B$ \cite{touvron2023llama}) remained frozen as well. The training remained on the parameters in alignment layer. The experiments were conducted in parallel on two NVIDIA H100 GPUs. We set the batch size as 6 and a learning rate as 0.0001. In Multi-View selection, the size $k$ of Multi-View group was set as 3. We also empirically set $\tau_h$ as 70\% and $\tau_l$ as 40\%.

\subsection{Multi-View selection}
Our model selected images from CLIP embedding. We picked a subset of egocentric images to input VLM. A representative example of each screen type is shown in Fig. \ref{Multi-View}. We noted that the selected images complemented each other in terms of field of view and spatial features, demonstrating that CLIP was an effective feature extractor for evaluating similarity and selecting views. Moreover, we analyzed the CLIP generated features using t-distributed Stochastic Neighbor Embedding (t-SNE) \cite{van2008visualizing}, to visualize the distribution of typical Multi-View images. It is observed that not only the features from the same Multi-View group is complement to each other but also the features among screen types are distinctive to each other. Such observation lays a foundation of text generation and screen type identification.

\begin{figure*}[!t]
\centering
\includegraphics[width=1\textwidth]{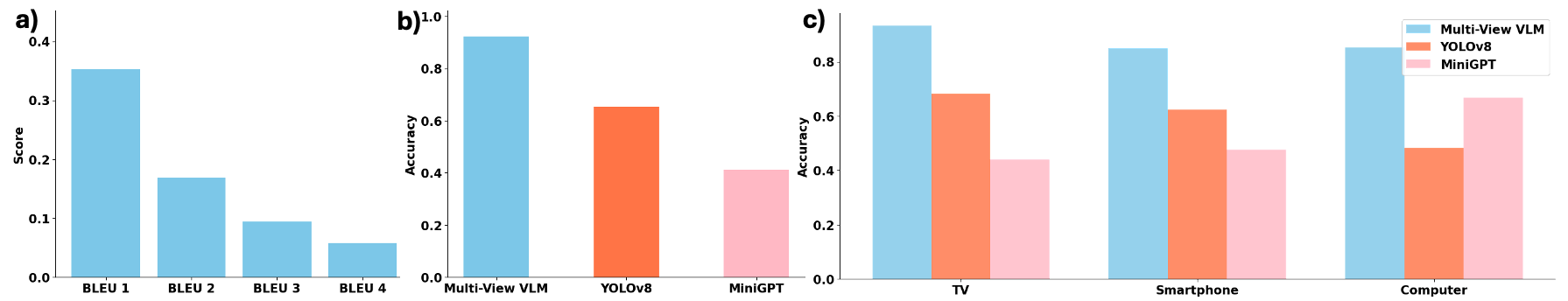}
\caption{a) Text generation results from proposed method (MV-VLM) and LLaVA to compare how close the generated scene description is similar to the ground truth annotation; b) Comparison among MV-VLM, CLIP-fC, LLaVA, YOLOv11, and MV-Swin-T in overall accuracy; c) Comparison among MV-VLM, CLIP-fC, LLaVA, YOLOv11, and MV-Swin-T in identification of TV, smartphone, and computer. Our methods consistently shows a higher performance than the other baseline methods.}
\label{fig:model_comparison}
\end{figure*}

\subsection{Text generation and screen type identification}
We evaluated the generated text description from Llama with the ground truth using BiLingual Evaluation Understudy (BLEU) \cite{papineni-etal-2002-bleu}. BLEU scores evaluate the similarity of the generated description by Llama and the reference caption. A higher BLEU indicates higher similarity to ground truth. BLEU $N$, where $N$ corresponds to contiguous sequences of $N$ word, indicates an evaluation of quality in a context. 
Moreover, our MV-VLM model was able to generated scene description with logic and smoothness.

Figure \ref{fig:text_res} shows representative scene descriptions in each screen type categories. Key words from each scene description is highlighted in red and categories to the three screen types (i.e., TV, smarpthone, and computer). ChatGPT4 is utlized to smooth the description for scene understanding. Our vision language model effectively identified various screen types, their spatial relationships, and action events with objects. For instance, in the first row, our model identified a laptop when the laptop screen is partially occluded. In this group, the light among the images shows high variance where the first two images are bright while the third image is dark.  In the second row, our proposed method identified a typical TV in a living room, even the screen is contorted through a plastic bottle. In the final row, when images are captured during motion, our proposed MV-VLM distinguished the smartphone from the environment in the blurry images.

\subsection{Comparison with existing methods}
To validate the superiority of MV-VLM over existing methods, we compared the performance of identifying screen types with a transformer based model (Multi-View Swin Transformer) \cite{sarker2024mv}, a convolution neural network (CNN) based model (YOLOv11) \cite{khanam2024yolov11overviewkeyarchitectural}, a multi-modal VLM (LLaVa) \cite{liu2023visualinstructiontuning}, and a modification of CLIP for classification task. The results are reported in Fig. \ref{fig:model_comparison}. The Multi-View Swin Transformer (MV-Swin-T) is a swin-transformer based algorithm that takes multiple images as input and classifies the input images. YOLOv11 is an object detection approach that uses CNN as backbone to output the bounding box and screen type. It is currently the state of the art that provides high accuracy in object detection. 
We concatenated the Multi-View images for the YOLOv11 as inputs. LLaVa is a multi-modal model that takes image and text inputs for caption generation.
CLIP for Classification (CLIP-fC) is a modification of CLIP by adding a fully-connected layer on top of CLIP features, enabling it for classification tasks. 
As shown in Fig. \ref{fig:model_comparison}, our MV-VLM outperformed the existing state-of-the-art (SOAT) algorithms.
Moreover, as shown in Fig. \ref{fig:text_res}, MV-LVM identified spatial relationship among objects (e.g., upon a wooden desk) and action events (e.g., holding a smartphone).

LLaVa took image and text as input and generate captions followed by the same key word extraction and identification process. The performance of LLaVa is much worse (around 15\%) than proposed method, indicating that it is necessary to retrain a VLM model with alignment layer specifically for screen type identification. The superiority of our method also demonstrates the need for Multi-View  processing in VLMs as LLaVa only processes images from single-view. Moreover, as shown in the text generation section in Fig. \ref{fig:model_comparison}, our model shows better in terms of BLEU scores compared to the LLaVA, which indicates the impotence of finetuning of VLM in the task of screen identification. The BLEU scores are only available for the VLM models such as LLaVA and MV-VLM.

The performance of MV-Swin-T and YOLOv11 are much worse (around 28\% and 25\%) compared to our algorithm in terms of average accuracy. These two models lack the language model components, which is necessary in this task to achieve optimal performance.
CLIP-fC unitizes a pretrained language model and the information from Multi-View images through major voting to acquire the classification results. It outperforms Multi-View Swin Transformer, YOLOv11, and LLaVa, but has worse average accuracy performance (around 9\%) compared to our method. The VLM-based models (MV-LVM, LLaVA, and CLIP-fC) out performs non-VLM models, which echoes the necessity of using VLMs for this application.

\subsection{Ablation study}

We conducted an ablation study to investigate the contributions of components in model architecture and view selection strategy. The results are reported in Table \ref{tab:ablation_study}.
To validate our model design, we separately removed four components from our design, which are LLM, Multi-View inputs, MiniLM, and Swin-Transformer. To remove the LLM in our framework, we employed a fully-connected layer to perform classification of screens based on the visual and text embeddings. Without Multi-View input, we use a single image and its corresponding environment description for text generation and screen type identification. To remove the MiniLM module, we wrapped the environment descriptions and visual embeddings to prompt and directly sent them to the alignment layer. To validate Swin-Transformer in our framework, we used ResNet for extracting the visual features. The results are reported in \textit{Model Archicture Ablation} in Table \ref{tab:ablation_study}. As shown in the table, our design outperforms other ablation variants, which echoes the necessity of the LLM, Multi-View image inputs, and Swin-Transformer in the framework.

\begin{table}[t]
\centering
\caption{Performance comparison of model architecture ablation and view selection strategy ablation variants across different screen types. The best performance is highlighted by \textcolor{red}{red} and the second best is highlighted by \textcolor{blue}{\uline{blue}}}

\renewcommand{\arraystretch}{1.5} 
\scalebox{0.8}{
\setlength{\tabcolsep}{0.2pt} 
\begin{tabular}{p{3cm}|ccccc}
\hline
\textbf{Ablation Variant} & \textbf{TV} & \textbf{Smartphone} & \textbf{Laptop} & \textbf{Non-screen} & \textbf{Acc.}   \\
\hline
\textbf{Ours}                        & \textbf{\textcolor{red}{0.9013}} & \underline{\textcolor{blue}{0.8753}} & \underline{\textcolor{blue}{0.9240}} & \textbf{\textcolor{red}{0.9561}} & \textbf{\textcolor{red}{0.9300}} \\
\hline
\multicolumn{6}{c}{\textit{Model Architecture Ablations}} \\
\textit{w/o LLM (use fully-connected layer)}  & 0.8944 & 0.8660 & 0.8948 & 0.8417 & 0.8650 \\
\textit{w/o Multi-View (use single image)}    & 0.8651 & \textbf{\textcolor{red}{0.8813}} & 0.9198 & 0.8972 & 0.8917 \\
\textit{w/o MiniLM (use direct input text)}   & 0.8614     & 0.8521     & \textbf{\textcolor{red}{0.9301}}     & 0.9218     & 0.8967   \\
\textit{w/o Swin-Transformer (use ResNet)}    & 0.7039  & 0.8063 & 0.8615 & 0.7077 & 0.7433 \\
\hline
\multicolumn{6}{c}{\textit{View Selection Strategy Ablations}} \\
\textit{Remove CLIP}                      & 0.8647     & 0.8515     & 0.8746     & \underline{\textcolor{blue}{0.9525}}     & \underline{\textcolor{blue}{0.9083}} \\
\textit{Replace CLIP with ResNet}                      & 0.8661  & 0.8435 & 0.8777 & 0.9447 & 0.9033 \\
\textit{Replace CLIP with KMeans}                      & \underline{\textcolor{blue}{0.8857}} & 0.8470 & 0.9027 & 0.9299  & 0.9050 \\
\hline
\end{tabular}}
\label{tab:ablation_study}
\end{table}

Moreover, we performed the ablation study on view selection strategies. Without the CLIP for view selection, we explored the following alternative options: 1) used temporally consecutive frames, 2) replaced CLIP with ResNet in view selection, and 3) replaced CLIP with KMeans in view selection. The results are reported in \textit{View Selection Strategy} section in Table \ref{tab:ablation_study}. As shown in the table, our Multi-View section design using CLIP embeddings achieves optimal performance.





\subsection{Hyperparameter search of $\tau_h$ and $\tau_l$}
\begin{figure}[!t]
\centering
\includegraphics[width=0.5\textwidth]{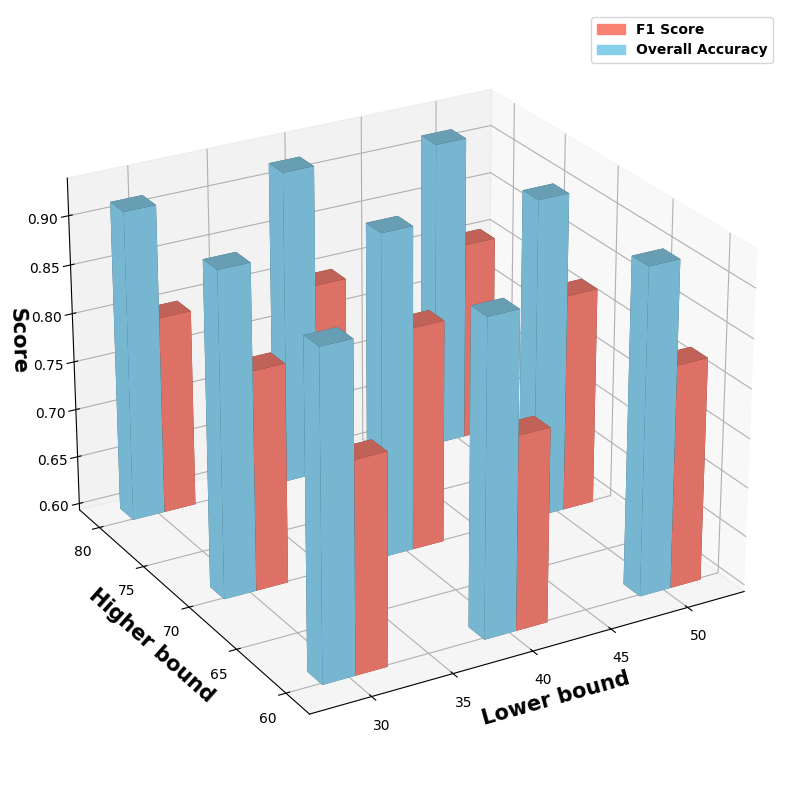}
\caption{The overall accuracy and F1 scores of hyperparameter search of $\tau_h$ and $\tau_l$.}
\label{fig:parameter_search}
\end{figure}

In our view selection algorithm, the choice of upper bound $\tau_h$ and lower bound $\tau_l$ determine the outcome of Multi-View frames. We performed a parameter search to identify the $\tau_h$ and $\tau_l$. As shown in Fig. \ref{fig:parameter_search}, the choice of $\tau_h$ as 70\% and $\tau_l$ as 40\% leads to the best F1 value and the second best Average Accuracy over different types of screens. It is worth mentioning that our view selection algorithm demonstrates robustness over the $\tau_h$ and $\tau_l$. In the nine groups of $\tau_h$-$\tau_l$ spanning from $30\%$ to $80\%$, the average accuracy scores are higher than 0.91 and the average F1 scores are higher than 0.79.

\subsection{Screen vs non-screen classification}
In behavioral study,  it is more important to determine the existence of electronic screens rather than identifying the screen types. To this end, we conducted a binary classification test to evaluate the performance of the MV-VLM in detecting the existence of screens or not. We had 600 groups of Multi-View images in the test set. The results, including a confusion matrix, are presented in Table \ref{binary}. The model achieved an overall accuracy of 95.5\% and a sensitivity of 93.67\%, indicating a stronger capability in detecting the screen existence with a small fraction of false negative detection. 

\begin{table}[t]
\centering
\caption{Confusion matrix from binary classification. We combine the classes from any type of screen as "screen" and the rest as "non-screen".}
\renewcommand{\arraystretch}{1.25}
\scalebox{0.8}{
\begin{tabular}{cccc}
\hline
                                                                             &                       & \multicolumn{2}{c}{Predicted Class}          \\ \cline{3-4} 
                                                                             &                       & Positive (screen) & Negative (non) \\ \hline
\multirow{2}{*}{\begin{tabular}[c]{@{}c@{}}Actual \\ Classes\end{tabular}} & Positive (screen)     & 281               & 19                     \\
                                                                             & Negative (non) & 8                & 292                   \\ \hline
\end{tabular}}
\label{binary}
\end{table}

\subsection{Assessment of comfort on wearable device}
We conduct an assessment of comfort on the wearable device. As shown in Table \ref{comfort}, the device receives high evaluation scores in movement, and emotional response, anxiety, and harm. The results suggest that the device is comfortable and acceptable for regular use by children. The high scores demonstrate the effectiveness of our tailored design for children, which includes light weight, a magnetic clip mount for secure attachment, a smaller and lighter badge-like design, and customized shapes to appeal to children. 

\begin{table}[t]
\centering
\caption{Assessment of Comfortness (All values are out of 10)}
\scalebox{0.8}{
\begin{tblr}{ 
}
\hline
\textbf{Metrics}                                             & \textbf{Mean} &  &  \\ \hline
Emotion \\ (Is it acceptable for children to wear the sensor?)            & 8.23                 &  &  \\
Harm \\ (The sensor does not cause pain or tickling.)                   & 8.36                 &  &  \\ 
Movement \\ (The sensor doesn't restrict children's moving.)  & 9.23                 &  &  \\
Anxiety \\ (Children feel secure wearing the device.)                   & 8.2                  &  &  \\ \hline 
\end{tblr}}
\label{comfort}
\end{table}


